\documentclass[runningheads]{llncs}

\usepackage[T1]{fontenc}
\usepackage{graphicx}
\usepackage{url}
\usepackage{booktabs}
\usepackage{array}

\title{Executable World Models for ARC-AGI-3 in the Era of Coding Agents}
\titlerunning{Executable World Models for ARC-AGI-3}

\author{Sergey Rodionov}
\authorrunning{S. Rodionov}
\institute{SingularityNET\\
\email{sergey@singularitynet.io}}

\begin{document}
\maketitle

\begin{abstract}
We evaluate an initial coding-agent system for ARC-AGI-3 in which the agent maintains an executable Python world model, verifies it against previous observations, refactors it toward simpler abstractions as a practical proxy for an MDL-like simplicity bias, and plans through the model before acting. The system is intentionally direct: it uses a scripted controller, predefined world-model interfaces, verifier programs, and a plan executor, but no hand-coded game-specific logic. The agent-facing prompts, workspace, and controller contain no game-specific code, game-specific prompts, hand-coded heuristics, hidden solutions, or other game-specific information; the same agent and prompts are used across games.

Because the coding agent has broad system access, we audit unintended information channels, describe earlier vulnerable harnesses, and explain how the current harness closes observed leakage channels while reducing benchmark-specific information exposure.

We report results on the 25 public ARC-AGI-3 games. Each playthrough starts from a fresh agent instance and clean workspace, with no access to files or conversation state from earlier playthroughs. With \texttt{GPT-5.5} high reasoning effort, the agent fully solved 15 games and achieved a mean per-game RHAE of 58.12\%. With \texttt{GPT-5.4} high reasoning effort, it fully solved 8 games and achieved a mean per-game RHAE of 41.29\%. Performance on the private validation set, which is not yet available to us, remains to be tested. Overall, the results provide preliminary evidence that verifier-driven executable world models are a promising approach for ARC-AGI-3 agents. Full run artifacts are released with the code at \url{https://github.com/astroseger/arc-3-agents-baseline1}.
\end{abstract}

\section{Introduction}

Large language models are most reliable when they are not used as final
authorities, but as proposal mechanisms inside systems that can check
their outputs. This pattern appears in recent successes such as
AlphaCode, where generated programs are filtered by execution
\cite{li2022alphacode}, FunSearch, where proposed programs are selected
by automated evaluators \cite{romeraparedes2024funsearch}, and
AlphaGeometry, where a neural model proposes auxiliary geometric
constructions and an exact symbolic engine derives and verifies their
consequences \cite{trinh2024alphageometry}. In each case, the language
model supplies approximate search, while reliability comes from an
external verification process.

This generate-and-verify pattern is powerful in domains such as
programming, algorithm discovery, and formal mathematics because
verification is relatively cheap. A program can be run against tests; a
candidate algorithm can be scored by an evaluator; a formal proof can be
checked by a proof assistant. General interactive agents face a different
problem. In an unfamiliar environment, each action may consume time,
change the state irreversibly, or reduce the final score. The agent
therefore cannot rely on trial and error in the environment as its main
source of verification. Instead, it needs an internal model in which
possible actions can be simulated, rejected, revised, and planned before
they are executed.

We use the term \emph{world model} in its standard broad sense: an
internal model that allows an agent to predict consequences, evaluate
hypotheses, and plan before acting. In this paper, the world model is not
a latent neural state or an opaque learned simulator. It is an
\emph{executable world model}: a Python codebase whose functions encode
the agent's current hypothesis about the environment. The model can be
run, tested, edited, and used for planning. It can also be refactored as
new observations arrive, so that accidental special cases are replaced by
simpler abstractions.

Our approach builds on prior work on world models and model-based agents.
World models are central to model-based reinforcement learning and agentic
intelligence more generally \cite{ha2018worldmodels}. Recent work has
also studied world models represented explicitly as programs. WorldCoder
builds a Python program representing an agent's knowledge of the world
from interaction data \cite{tang2024worldcoder}. Code World Models study
world models generated by large language models in the form of Python code
for model-based reinforcement learning \cite{dainese2024codeworldmodels}.
The contribution of this work lies in adapting this programmatic
world-modeling perspective to ARC-AGI-3 and studying how a coding agent
can maintain, verify, refactor, and use such a model under strict
interaction constraints.

A second principle is simplicity. When an agent observes only a small number of transitions, many world models may be consistent with the data. The useful model is not merely one that fits previous observations, but one that captures the underlying regularities compactly enough to support future planning. This is closely related to the Minimum Description Length perspective, which treats good explanations as those that compress the observations while accounting for the complexity of the model itself \cite{grunwald2004mdl}. In our system, however, we do not implement a formal MDL objective. Instead, we use a practical proxy suited to coding agents: after new observations are incorporated, the agent is repeatedly asked to refactor its executable model, replacing special cases with simpler abstractions while preserving verifier correctness. This connects to prior work on library learning and code refactoring, where reusable abstractions and MDL-like compression have been used to guide or evaluate better code organization \cite{ellis2020dreamcoder,grand2023lilo,kovacic2025refactoring}.

ARC-AGI-3 is a natural testbed for this approach. Unlike static puzzle benchmarks, ARC-AGI-3 places agents in novel, abstract, turn-based environments where they must explore, infer goals, build models of environment dynamics, and plan action sequences without explicit natural-language instructions \cite{arcprize2026arcagi3}. The benchmark is explicitly framed around adaptive efficiency: humans can solve the environments, while frontier AI systems scored below 1\% as of March 2026 \cite{arcprize2026arcagi3}. This makes ARC-AGI-3 a controlled version of a broader AGI problem. The environments are simple enough that explicit computational models are plausible, but direct experimentation is still costly enough that blind trial and error is inadequate. To perform well, an agent must learn from limited interaction, compress observations into a useful model, and use that model to decide which actions are worth spending in the real environment.

In this paper, we study an initial agent based on this principle. The
agent is instructed to maintain a Python codebase representing its current
model of the game, including functions for state representation,
transition prediction, goal checking, and planning. After each
modification of the world model, the agent is instructed to run verifiers
that test consistency with previous observations. The agent is also
prompted to refactor the model as evidence accumulates, replacing special
cases with simpler executable abstractions.

We evaluate the agent under strict ARC-AGI-3 interaction constraints. The
unit of evaluation is a recorded playthrough: for each playthrough, a fresh
agent process and clean workspace are launched, and the agent receives only
one exposure to the target game. Within that playthrough, the agent may not
restart the whole game to obtain a better trajectory, may not return to
previously completed levels, may not use information from previous
playthroughs of the same game, and is not given hand-coded game-specific
logic.
We report the performance of our agent on all 25
public games. The agent and evaluation harness are designed to be game-general
within ARC-AGI-3 rather than tailored to individual games. In practice, most
of the development of this first version was performed using the public game
\texttt{ls20}. However, only evaluation on the private validation set
can directly test how well the approach generalizes. We report this first
implementation as a baseline for studying verifier-driven executable
world models in ARC-AGI-3.

\section{Implementation}
\label{sec:implementation}

\subsection{ARC-AGI-3 interaction protocol}

ARC-AGI-3 games are interactive, level-based environments. At each step,
the agent observes the current game state and submits one action from the
available action set. A level attempt may end in either
\texttt{LEVEL\_COMPLETED} or \texttt{GAME\_OVER}. If \texttt{GAME\_OVER}
is reached, the benchmark-provided \texttt{RESET} action can be used to
restart the current level attempt within the same game run. \texttt{RESET} is treated as an ordinary environment action
and counts against the action budget. The agent cannot restart the whole game to
obtain a better run, and the agent cannot return to previously completed
levels.

\subsection{Agent architecture}

Our architecture is intentionally simple. The system consists of a coding
agent controlled by a scripted external controller. The coding-agent
runtime in these experiments was Codex CLI version \texttt{0.128.0},
an OpenAI coding agent that can read, edit, and execute code in a local
working directory \cite{openai_codex_cli}. The external
controller does not solve levels directly. Instead, it starts the game,
passes observations to the coding agent, monitors whether the current
level is still running, completed, or in \texttt{GAME\_OVER}, and sends
predefined prompts in these situations.

During normal play, the coding agent is instructed to continue working
until it either completes the current level or reaches
\texttt{GAME\_OVER}. Before normal continuation and when progress appears to stall, the
controller asks the coding agent to simplify and refactor its world model
before continuing. If \texttt{GAME\_OVER} is reached, the controller
issues \texttt{RESET}, returns the new attempt observation, and asks
the coding agent to simplify and refactor the world model before
continuing.

The initialized Python workspace contains templates for transition
dynamics, state reconstruction and rendering, and planning. In the public
implementation, these correspond to \texttt{world\_model\_engine.py},
\texttt{world\_model\_state\_io.py}, and
\texttt{world\_model\_main\_planner.py}. These files initially contain
only predefined interfaces; the coding agent is responsible for filling
them in and maintaining them as new levels and observations are
encountered.

\subsection{Verification and execution helpers}

The workspace also provides general helper programs. The most important
are the world-model verifier, the planner verifier, and the plan
executor. The world-model verifier checks that the executable model
reproduces the recorded observations from previous attempts. The planner
verifier checks that the main planner can produce plans that reach
\texttt{LEVEL\_COMPLETED} inside the learned model for solved levels.
Planner-running utilities allow the agent to test plans from the current
state, from the initial state of a level, or from an intermediate point in
a previous attempt.

The plan executor is the main interface between planning and real
environment actions. Given a proposed sequence of actions, it simulates
the sequence in the world model and executes the same actions in the real
game. After each non-terminal step, it compares the predicted settled
ASCII frame with the observed settled ASCII frame. If the prediction
diverges from the observation, the executor stops immediately and records
mismatch artifacts for inspection. It also stops on
\texttt{LEVEL\_COMPLETED} or \texttt{GAME\_OVER}. Thus, successful plan
execution is not merely action replay: it is an online test of the
current world model.

In the current implementation, the coding agent still has direct access
to the game client and can therefore bypass the plan executor. The prompt
instructs the agent to use the plan executor as early as possible and to
treat any mismatch as a blocking modeling error.

\subsection{Refactoring loop}

The coding agent is repeatedly prompted to keep the world model compact
and general. In particular, it is asked to replace special cases with
shared rules, simplify state reconstruction, remove ad hoc rendering
overrides, and keep the planner expressed in terms of the world-model
engine. This refactoring loop is our practical proxy for an MDL-like
bias: the model should not merely fit the observed transitions, but
should explain them through a simpler executable structure that remains
valid as new levels are encountered.

\subsection{Runtime recovery}

Agent runs occasionally encounter failures in the coding-agent runtime.
In our experiments, Codex sometimes failed because of transient
service-side errors, such as capacity errors or other OpenAI-side
failures. In other cases, the Codex process appeared to freeze: it continued
running, but stopped making observable progress. These events were
relatively rare and did not occur in every game, but over a full 25-game
evaluation we usually observed a small number of such agent-runtime
interruptions.

The runner therefore includes a recovery mechanism for the
coding agent; this mechanism restarts only the agent process and does not
restart or modify the game server. If the agent exits unexpectedly, or if
the runner detects that Codex has stalled and terminates it, the runner
restarts the agent in recovery mode. In this mode, the agent receives a special prompt explaining that the
previous session was interrupted, followed by an additional refactoring
loop. The purpose of this recovery prompt is to
help the agent reconstruct the current state of the run from the
workspace before continuing. We allow at most ten such recoveries in a
single game run. This limit is intended to handle ordinary transient
agent-runtime failures, not long service outages; for example, a
multi-hour OpenAI-side outage would still cause the run to fail.

\subsection{Operating cost}

The system is relatively inexpensive to run. In our experiments with the
\texttt{GPT-5.5} high-reasoning configuration, a single ChatGPT Pro
subscription, priced at USD 200 per month at the time of writing, was
sufficient to run full experiments for roughly two to eight games,
depending on game difficulty, within the weekly Codex usage limit for
that subscription. This estimate is only meant to give the
approximate scale of the cost, since subscription limits and model availability may change
over time.

\section{Preventing unintended information access}
\label{sec:data-leakage-controls}

Because our agent has broad system access, including the ability to inspect files and execute code, we took additional steps to prevent it from accessing information outside the intended game interface. In other words, we wanted to ensure that the agent could not exploit unintended information channels. For example, the agent should not be able to identify the real game name, or discover that the environment is connected to ARC or ARC-AGI, because such identifiers could become exploitable if any external lookup channel were available. Those lookup channels should also be unavailable: the agent should have no general internet access. Similarly, the agent should not be able to run a second game client to test moves outside the official interaction protocol.

Early versions of our evaluation harness already included several countermeasures. The agent was run inside an isolated Docker environment, the ARC-AGI-3 library was hidden behind a client--server interface, and the prompt did not reveal the real game name or the terms \texttt{ARC} or \texttt{ARC-AGI}. However, later audits exposed several residual leakage channels. Some client-side files still contained comments or environment variables containing \texttt{ARC}. The true game identifier was not shown in the prompt, but it could still be recovered indirectly. For example, the server API exposed it, even though the client filtered this information and did not send it to the agent directly. In addition, the external controller was started with the real game name as a process argument, so the agent could recover it by inspecting running processes, e.g. with \texttt{ps}. Finally, the agent container had internet access, and Codex web search was enabled.

One interesting observation is that the agent was sometimes able to exploit these leakage channels. These weaknesses became especially visible during ablation experiments. Simpler agents, with fewer constraints on their behavior, were more likely to use unintended channels. We observed two main classes of undesirable behavior. First, an agent could infer or recover the real game identifier and then use web search to look for public scorecards or solution traces. Second, when \texttt{arc\_agi} was run in local, non-competition mode, an agent could start a second game client and use it as an unscored simulator for trying actions. This attack is not available in the competition setting, where the \texttt{arc\_agi} library does not allow the same game to be started a second time, but it was possible in our local development setup.

We audited historical runs for these behaviors. In an earlier vulnerable version of the evaluation harness, we found one clear type-1 leakage case: a previous \texttt{GPT-5.5}-medium run on \texttt{dc22} appears to have downloaded a scorecard or related external material. This run is retained in the public repository for transparency, but we do not use any results computed with the old vulnerable version in this article, except when explicitly discussing examples of leakage. In an even older \texttt{GPT-5.4} run, the agent appears to have started a parallel game client. We also found several cases where the agent attempted to use web search but did not appear to obtain useful external information. Thus, apart from one clear case, we did not find strong evidence that the old agent substantially benefited from these leakage channels. Nevertheless, we discard these old results and do not use them in the article unless explicitly stated.

The current evaluation harness closes the observed leakage channels. The agent container no longer contains the real game name in files, process arguments, environment variables, or API-visible services. It also no longer contains references to \texttt{ARC} or \texttt{ARC-AGI}. This reduces both the direct lookup risk and the risk that the model could exploit memorized information about public ARC-AGI-3 games, since the agent does not know which benchmark or game it is interacting with. The agent container has no general internet access. It can only reach OpenAI services through a separate proxy container whose allowlist is restricted to OpenAI endpoints. Codex web search is disabled. Finally, the client--server protocol has been changed so that starting a second game client is rejected even in local mode.

These controls do not constitute a formal proof of security against an arbitrarily capable adversarial agent. For example, a sufficiently capable coding agent might try to modify its own configuration or source code to re-enable disabled functionality. However, such behavior has not been observed in our current runs. Moreover, even if web search were somehow re-enabled, the agent would still lack the real game identifier and would not be told that the task is related to ARC-AGI-3. Overall, we believe that the current evaluation harness prevents the cheating behaviors observed in earlier versions. In addition, by removing any accessible game identifier and any indication that the task is related to ARC-AGI-3, these controls reduce the risk that the agent could exploit memorized information about public games from the model weights.

\section{Results on 25 Public Games}
\label{sec:results-public}

We evaluated the agent on all 25 public ARC-AGI-3 games. We report two
sets of runs: \texttt{GPT-5.5} with high reasoning effort and
\texttt{GPT-5.4} with high reasoning effort.

Each game is evaluated as a single recorded playthrough. The agent starts
from a fresh workspace, sees the target game only once, and has no access
to previous playthrough-specific files, logs, or conversation state. It
cannot restart the whole game to obtain a better trajectory, and it
cannot return to previously completed levels. We impose a per-level cap
of 1500 environment actions. When the cap is exceeded on an unsolved
level, the playthrough is stopped and the score reached so far is
reported. Because actions are submitted by the coding agent while the
stopping rule is enforced by the external controller, the number of actions recorded on
the final unsolved level can slightly exceed this cap.

\paragraph{Interrupted runs.}
As described in Section~\ref{sec:implementation}, the runner can recover
from ordinary transient failures of the coding-agent runtime. However,
some interruptions cannot be recovered, for example prolonged
OpenAI-side outages or Codex usage limits on the OpenAI
subscription used for the experiment. We report these runs as interrupted
and do not restart them. Restarting after an interruption would conflict with the
single-playthrough evaluation protocol. It would also introduce a
selection bias. For the same game, runs in which the agent struggles and
obtains a lower score usually take longer than runs in which it finds a
good solution quickly. Longer runs have more opportunity to encounter a
technical interruption or hit a Codex usage limit. Restarting after
such failures would therefore preferentially discard some low-scoring
trajectories.

\paragraph{Scoring.}
We report Relative Human Action Efficiency (RHAE), the official
efficiency-based score for ARC-AGI-3 \cite{arcprize2026arcagi3}. We also
report the number of levels solved, the run status, and two action
counts. In the tables, $A_{\mathrm{solved}}$ is the number of
environment actions spent on solved levels, and
$A_{\mathrm{final}}$ is the number of environment actions spent on the
final unsolved level, if any.

\paragraph{Local execution.}
All runs reported in this section used the ARC-AGI-3 library in local mode, while
the evaluation harness enforced the competition interaction rules. We did not use competition
scorecard mode directly because it currently has operational time limits:
the scorecard closes after 15 minutes of inactivity and after 24 hours in
total. These time limits are operational constraints of the current
competition mode, not part of the ARC-AGI-3 evaluation protocol itself.
They are incompatible with coding-agent runs that may spend more than 15
minutes in a refactoring loop and may take up to roughly 48 hours on
difficult games. In practice, we run several games sequentially on each
OpenAI subscription. Running all 25 games within a single 24-hour
scorecard window would therefore require more OpenAI subscriptions than
we currently have, so a full evaluation usually takes several days.

Local mode has one behavior that is not allowed by our evaluation
protocol: sending \texttt{RESET} before taking any action on a level can
restart the whole game from the first level. The coding agent itself is
not instructed to use \texttt{RESET}; this action is normally sent by
the external controller after \texttt{GAME\_OVER}. Occasionally, the
coding agent nevertheless discovers that it can send \texttt{RESET}
itself, which we do not treat as problematic by itself. However, sending
\texttt{RESET} before taking any action on a level can trigger a
full-game restart. If it occurs, we treat the restart as terminal for
evaluation purposes and discard all actions taken after the reset when
computing the reported score. This situation did occur in one reported run, \texttt{s5i5} with \texttt{GPT-5.4}.

\begin{table}[t]
\centering
\caption{Public-game results with \texttt{GPT-5.4} high reasoning effort.}
\label{tab:public-results-gpt54}
\scriptsize
\begin{tabular}{lrrrrl}
\hline
Game & Levels & RHAE & $A_{\mathrm{solved}}$ & $A_{\mathrm{final}}$ & Status \\
\hline
\texttt{ar25} & 8/8 & 92.80\% & 521 & -- & normal termination \\
\texttt{bp35} & 3/9 & 2.73\% & 481 & 1199 & interrupted \\
\texttt{cd82} & 6/6 & 100.00\% & 147 & -- & normal termination \\
\texttt{cn04} & 1/6 & 0.33\% & 111 & 1514 & normal termination \\
\texttt{dc22} & 4/6 & 36.84\% & 1568 & 715 & interrupted \\
\texttt{ft09} & 6/6 & 100.00\% & 109 & -- & normal termination \\
\texttt{g50t} & 6/7 & 59.25\% & 860 & 785 & interrupted \\
\texttt{ka59} & 6/7 & 9.29\% & 2220 & 161 & interrupted \\
\texttt{lf52} & 1/10 & 1.82\% & 10 & 1511 & normal termination \\
\texttt{lp85} & 8/8 & 100.00\% & 110 & -- & normal termination \\
\texttt{ls20} & 7/7 & 77.26\% & 804 & -- & normal termination \\
\texttt{m0r0} & 1/6 & 0.01\% & 790 & 1519 & normal termination \\
\texttt{r11l} & 4/6 & 18.45\% & 693 & 502 & interrupted \\
\texttt{re86} & 4/8 & 27.78\% & 187 & 1390 & interrupted \\
\texttt{s5i5} & 5/8 & 41.67\% & 297 & 423 & normal termination \\
\texttt{sb26} & 8/8 & 73.87\% & 234 & -- & normal termination \\
\texttt{sc25} & 3/6 & 16.71\% & 1059 & 740 & interrupted \\
\texttt{sk48} & 1/8 & 2.78\% & 30 & 1163 & interrupted \\
\texttt{sp80} & 1/6 & 4.76\% & 39 & 1479 & interrupted \\
\texttt{su15} & 1/9 & 1.72\% & 25 & 1582 & normal termination \\
\texttt{tn36} & 1/7 & 3.57\% & 11 & 1549 & normal termination \\
\texttt{tr87} & 6/6 & 100.00\% & 240 & -- & normal termination \\
\texttt{tu93} & 9/9 & 100.00\% & 191 & -- & normal termination \\
\texttt{vc33} & 3/7 & 10.73\% & 132 & 1529 & normal termination \\
\texttt{wa30} & 7/9 & 49.77\% & 1290 & 919 & interrupted \\
\hline
\end{tabular}
\end{table}

With \texttt{GPT-5.4}, the agent fully solved 8 of 25 games and achieved
a mean per-game RHAE of 41.29\%.

\begin{table}[t]
\centering
\caption{Public-game results with \texttt{GPT-5.5} high reasoning effort.}
\label{tab:public-results-gpt55}
\scriptsize
\begin{tabular}{lrrrrl}
\hline
Game & Levels & RHAE & $A_{\mathrm{solved}}$ & $A_{\mathrm{final}}$ & Status \\
\hline
\texttt{ar25} & 8/8 & 100.00\% & 307 & -- & normal termination \\
\texttt{bp35} & 5/9 & 4.43\% & 1195 & 1340 & interrupted \\
\texttt{cd82} & 6/6 & 92.91\% & 170 & -- & normal termination \\
\texttt{cn04} & 6/6 & 96.34\% & 715 & -- & normal termination \\
\texttt{dc22} & 0/6 & 0.00\% & 0 & 1586 & normal termination \\
\texttt{ft09} & 6/6 & 57.80\% & 474 & -- & normal termination \\
\texttt{g50t} & 7/7 & 95.08\% & 556 & -- & normal termination \\
\texttt{ka59} & 7/7 & 100.00\% & 541 & -- & normal termination \\
\texttt{lf52} & 6/10 & 35.48\% & 632 & 1580 & normal termination \\
\texttt{lp85} & 8/8 & 100.00\% & 190 & -- & normal termination \\
\texttt{ls20} & 7/7 & 57.02\% & 1600 & -- & normal termination \\
\texttt{m0r0} & 6/6 & 67.33\% & 2441 & -- & normal termination \\
\texttt{r11l} & 6/6 & 89.29\% & 182 & -- & normal termination \\
\texttt{re86} & 5/8 & 40.10\% & 423 & 272 & interrupted \\
\texttt{s5i5} & 2/8 & 0.25\% & 860 & 18 & interrupted \\
\texttt{sb26} & 8/8 & 83.78\% & 234 & -- & normal termination \\
\texttt{sc25} & 3/6 & 20.23\% & 92 & 347 & interrupted \\
\texttt{sk48} & 7/8 & 11.26\% & 3496 & 0 & interrupted \\
\texttt{sp80} & 4/6 & 39.21\% & 178 & 195 & interrupted \\
\texttt{su15} & 9/9 & 56.85\% & 736 & -- & normal termination \\
\texttt{tn36} & 7/7 & 74.98\% & 606 & -- & normal termination \\
\texttt{tr87} & 6/6 & 100.00\% & 416 & -- & normal termination \\
\texttt{tu93} & 9/9 & 100.00\% & 266 & -- & normal termination \\
\texttt{vc33} & 3/7 & 21.43\% & 61 & 1325 & interrupted \\
\texttt{wa30} & 4/9 & 9.11\% & 1186 & 1511 & normal termination \\
\hline
\end{tabular}
\end{table}

With \texttt{GPT-5.5}, the agent fully solved 15 of 25 games and achieved
a mean per-game RHAE of 58.12\%.

Performance remains uneven and can vary substantially across runs.
Although \texttt{GPT-5.5} is stronger in aggregate,
\texttt{GPT-5.4} obtained higher scores on some games. On
\texttt{ft09}, it achieved 100.00\% RHAE, compared with 57.80\% for
\texttt{GPT-5.5}. On \texttt{wa30}, it solved 7/9 levels with
49.77\% RHAE, compared with 4/9 levels and 9.11\% RHAE for
\texttt{GPT-5.5}. Conversely, \texttt{GPT-5.5} solved all 7 levels of
\texttt{tn36} with 74.98\% RHAE, which was much higher than we usually
observed for this game in earlier experiments using vulnerable versions
of the evaluation harness. Those earlier results are not included in the
reported scores, but they also showed substantial variation between runs.
Overall, these cases should be read as evidence of run-to-run
variability, rather than as precise per-game comparisons between the two
models.

\section{Discussion}
\label{sec:discussion}

The public-game results are encouraging for a first implementation. The
agent was assembled with a relatively simple architecture: a coding agent,
a fixed external controller, executable world-model templates, verifiers,
and a plan executor. Nevertheless, with \texttt{GPT-5.5} as the
underlying language model, it fully solved 15 of the 25 public games and
achieved a mean per-game RHAE of 58.12\%. This suggests that
verifier-driven executable world models are a promising approach for
ARC-AGI-3 agents.

At the same time, executable world models alone do not make the system
reliable. Performance remains uneven across games and can vary
substantially across runs. Some environments are modeled and exploited
effectively, while others lead to near-total failure. This variability is
especially important if the goal is to approach 100\% score: for example,
the system solved \texttt{ft09} with 100.00\% RHAE in the
\texttt{GPT-5.4} run, but obtained only 57.80\% RHAE on the same game in
the \texttt{GPT-5.5} run.

One possible explanation is premature commitment to an incorrect or overly
specific world model. Once the agent has built such a model, it may
continue refining and planning within it instead of actively considering
alternatives. This suggests several possible improvements: better
exploration prompts, explicit competing-hypothesis tracking, and prompting the agent
to choose actions whose outcomes would test or falsify its current model.

Another way to address this problem would be to maintain multiple world
models in parallel. For example, several agents could independently build
candidate models while only one agent is allowed to spend environment
actions. Another agent or controller could then select among models based
on model generality. A related design would keep one active agent but periodically ask it to
inspect, revise, or borrow ideas from inactive alternative models. These
multi-model protocols would add complexity, but they directly target the
hypothesized failure mode: committing too early to a single wrong model.

Another direction is to provide more reusable agentic skills. The current
system mostly relies on the coding agent to invent its own search,
planning, debugging, and model-comparison routines during each run. Future
versions could include reusable routines for breadth-first search, A*
search, symbolic constraint solving, backtracking, subgoal decomposition,
state abstraction, model comparison, and strategy selection. These skills
would not encode game-specific solutions, but would give the agent a
stronger general toolkit for using and improving its executable world
model.

From the ARC-AGI-3 perspective, the main lesson is that a coding agent can
sometimes use a compact executable model as an internal testbed for
planning. This is different from directly trial-and-erroring in the
environment. The agent spends unscored computation editing, testing, and
simulating the world model, and spends scored environment actions only on
selected plans. The current system is still inefficient and brittle, but
the successes indicate that the basic loop---observe, model, verify,
refactor, plan, execute---is viable on a nontrivial subset of public
games.

From the AGI perspective, these results should be treated as a limited
but useful case study. ARC-AGI-3 is far simpler than the real world: its
environments are discrete, deterministic, and visually abstract. However,
it captures an important structural problem for general agents. Direct
interaction is costly, and the agent must decide which hypotheses and
plans are worth testing. In such settings, world models play a role
analogous to verifiers in programming or formal reasoning: they provide
an internal space in which candidate actions can be evaluated before
acting. The open question is how to scale this idea from small,
hand-initialized executable models to richer, hierarchical models of
real-world domains.

\paragraph{Future directions.}
The current system is a first, relatively ad hoc implementation of the
main idea. A natural next step is an ablation study. Our preliminary
experiments suggest that simpler versions of the agent may achieve
comparable performance, but those experiments used vulnerable versions of
the evaluation harness and are therefore not included in the reported
results. A clean ablation study with the current harness would help
identify which components are most important and where improvement effort
should be focused.

The discussion above suggests several possible improvements: better
hypothesis management, multi-model protocols, and a stronger library of
reusable agentic skills. These changes target the main apparent weakness
of the current system: it can build useful executable models, but it is
not yet reliable in how it explores, compares alternatives, and uses those
models.

A broader direction is to adapt the same architecture beyond ARC-AGI-3.
With relatively small changes, the agent could be turned into a general
world-modeling agent for tasks specified by a prompt in low-dimensional,
deterministic environments. We have preliminary versions of such a system;
the next step is to evaluate it across a wider range of environments.

Overall, the present system should be viewed as a baseline rather than a
finished architecture. The private validation set is the decisive test of
whether the current ARC-AGI-3-specific tools are genuinely game-general
within the benchmark or whether they have indirectly adapted to the
public games.

\section{Conclusion}

We presented an initial ARC-AGI-3 agent that uses a coding agent to build
an executable Python world model, verify it against observed transitions,
refactor it toward simpler abstractions, and plan through it before
spending environment actions. On the 25 public ARC-AGI-3 games, the
\texttt{GPT-5.5} version fully solved 15 games and achieved a mean
per-game RHAE of 58.12\%, while the \texttt{GPT-5.4} version fully solved
8 games and achieved a mean per-game RHAE of 41.29\%. These results
provide preliminary evidence that verifier-driven executable world models
are a promising approach for ARC-AGI-3 agents.

At the same time, the system remains a baseline rather than a finished
architecture. Performance is uneven, varies across runs, and still
depends heavily on how the agent explores, forms hypotheses, and uses its
world model. The most important next steps are clean ablations, better
hypothesis management, adding reusable agentic skills, and evaluation on
private validation games once they are available.

\section*{Code Availability}
The project repository is available at:

\noindent\url{https://github.com/astroseger/arc-3-agents-baseline1}

\begin{credits}
\subsubsection{\discintname}
The author has no competing interests to declare that are relevant to the content of this article.
\end{credits}

\bibliographystyle{splncs04}
\bibliography{arc_agi3_article1_references}

\end{document}